# TAVI-TEC: An AI-Based Tool for Procedural Planning of Transcatheter Aortic Valve Implantation

Alessandra Zerillo[1], Stefano Cannata[2], Diego Bellavia[1], Daniele Ciriello[5], Simone Manini[3,5], Salvatore Pasta[1,4], Caterina Gandolfo[2]

[1] Department of Research, IRCCS-ISMETT, Palermo, Italy.

[2] Department for the Treatment and Study of Cardiothoracic Diseases and Cardiothoracic Transplantation, IRCCS-ISMETT, Palermo, Italy.

[3] D/Vision Lab, Dalmine, Italy.

[4] Department of Engineering, Viale delle Scienze, Università degli Studi di Palermo, Palermo, Italy.

[5] DICOM Vision, Dalmine, Italy

*** Correspondence:**

Salvatore Pasta, PhD
Professor of Industrial Bioengineering
Department of Engineering
University of Palermo
e-mail: salvatore.pasta@unipa.it
and
Department of Research,
IRCCS Mediterranean Institute for Transplantation and Advanced Specialized Therapies
Via Tricomi 5, Palermo, Italy
e-mail: spasta@ismett.edu

**Abstract**

Computed tomography angiography (CTA) is crucial for preprocedural TAVI planning, providing the anatomical information required for prosthesis sizing and vascular access assessment. As the volume of TAVI procedure increases, improving efficiency and standardizing annotations is becoming essential in clinical practice. This study presents TAVI-TEC, a fully automated artificial intelligence-based framework integrated into a web-based DICOM viewer for routine preoperative TAVI planning. Pre-procedural CTA scans from patients undergoing TAVI with SAPIEN 3 Ultra (S3U) prostheses were processed using a fully automated pipeline. Deep learning-based segmentation of cardiovascular structures, calcification detection, centerline extraction, landmark identification, and annular plane definition was implemented to quantify key annular and aortic root measurements and color-coded maps of lumen reduction and vessel diameter for vascular access. A multilayer perceptron classifier was trained to predict prosthesis size prior to the TAVI procedure. Results revealed that TAVI-TEC enabled pre-procedural measurements in approximately 2–6 min. Strong agreement with clinician-derived measurements was observed for annular area (coefficient of concordance, CCC = 0.934; interclass correlation coefficient, ICC = 0.935; $R^2 = 0.881$) and perimeter (CCC = 0.909; ICC = 0.909; $R^2 = 0.854$). The valve-size prediction model achieved 82% overall accuracy, with most misclassifications occurring between adjacent prosthesis sizes. Though further multicenter validation and extension to additional measurements and valve platforms are required, the TAVI-TEC methodology may reduce operator variability in pre-TAVI measurements and streamline the preoperative workflows of the Heart Team for decision-making.

## 1. INTRODUCTION

Transcatheter aortic valve implantation (TAVI) has become a less invasive alternative to open-heart surgery for patients with severe aortic stenosis and is increasingly performed worldwide [1, 2]. Initially, TAVI was limited to patients at high surgical risk, but indications were progressively expanded to include patients at intermediate and low surgical risk [3, 4, 5]. The aortic annulus is a complex three-dimensional, oval-shaped structure with a conical configuration bounded at nadir by the attachments of the three aortic leaflets. Accurate geometric and dimensional measurement of this structure is essential for device selection and optimal bioprosthetic valve sizing [6, 7]. Valve sizing is primarily based on annular area for balloon-expandable prostheses and annular perimeter for self-expanding prostheses, with additional anatomical measurements of the aortic root. Accurate measurements are essential for selecting the most appropriate prosthesis for each patient in order to improve clinical outcomes and reduce complications. Specifically, undersizing may increase the risk of paravalvular leak and prosthesis–patient mismatch, whereas oversizing may increase the risk of annular rupture and conduction disturbances [8]. These complications are associated with increased morbidity and mortality following TAVI [9, 10].

Computed tomography angiography (CTA) plays a central role in the pre-operative planning of TAVI by providing the essential anatomical information and annotations required for valve sizing and vascular access assessment [11]. CTA enables evaluation of the aortic valve complex and valve calcification whereas full-body imaging supports assessment of the iliofemoral access route for vascular access of the predicated device. Manual annotation is needed to quantify pre-TAVI measurements from diagnostic imaging but can lead to intra-operator variability and requires additional annotations (eg, measurements of sinus height, coronary access) for clinical decision-making in patients with borderline anatomies.

Recent studies have developed fully or semi-automatic software tools to provide reproducible measurements of the annulus anatomical structures and reduce errors on anatomic measurements induced by interobserver variability [12-15]. A common feature of these applications is the initial segmentation of aortic root anatomy by deep learning followed by geometric analysis to extract anatomic measurements. Machine learning was also used for prediction of the optimal device sizing of the predicate device followed by validation in large patient cohort. However, some tools partially rely on manual input. For example, in 3Mensio, measurements are semi-automatic: the nadir of each cusp is suggested by the algorithm and can subsequently be adjusted by the operator when needed. Moreover, not all relevant parameters are evaluated, underscoring the need for fully automated solutions. Table 1 summarizes key research and funding of previous studies on automatic solutions for pre-TAVI measurements.

**Table 1**. Comparative overview of AI-based tools for pre-procedural TAVI assessment:

| | **Saitta et al.[12]** | **Krüger et al.[13]** | **Santaló-Corcoy et al.[15]** | **Toggweiler et al.[14]** |
|---|---|---|---|---|
| **Method** | 3D U-Net with differential geometry | Cascaded CNNs with PCA | MeshDeformNet with 3D residual U-Net | 2D/3D U-Nets with landmark detection |
| **Runtime** | <45 s | ~30 s | ~2 min | ~10 min |
| **Measurements** | Annulus, STJ, sinuses, calcium | Annular diameter and orientation | 22 measurements, including annulus, LVOT, SOV, and coronary ostia | Annulus, SOV, STJ, and ascending aorta |
| **Device-sizing prediction** | Not specified | 61–81% | Not specified | 87–88% |

**Note**: CNN, convolutional neural network; LVOT, left ventricular outflow tract; PCA, principal component analysis; SOV, sinus of Valsalva; STJ, sinotubular junction; U-Net, convolutional neural network architecture

This study presents TAVI-TEC, a fully automated pipeline that segments cardiac anatomy from CT scans, extracts key pre-TAVI anatomical measurements, and advises on the optimal prosthetic valve size in few minutes (≤ 7 min) without requiring user interaction. After testing TAVI-TEC on population study group of 109 TAVI patients, predictions of annulus dimensions were compared with manual measurements performed by an experienced cardiologist. Finally, the TAVI-TEC methodology was implemented as a dedicated module within the DICOM Vision web-based medical imaging platform developed by D/Vision Lab, enabling visualization, review, and management of AI-generated measurements within an imaging workflow. By providing consistent measurements and valve sizing in a time-efficient manner, TAVI-TEC is intended to support decision-making and streamline pre-procedural planning of TAVI patients.

## 2. METHODS

### 2.1 Patient Population

We included 109 consecutive patients with severe aortic stenosis who underwent transcatheter aortic valve implantation (TAVI) with the SAPIEN 3 Ultra (S3U) transcatheter heart valve at IRCCS ISMETT (Palermo,

Italy). Pre-procedural CTA was performed as part of routine clinical care using a 256-row dual-source CT scanner (SOMATOM; Siemens Healthineers). Pixel spacing ranged from 0.26 x 0.26 mm2 to 0.87 x 0.87 $mm^2$, and slice thickness was 0.25 mm. Chest and full-body imaging datasets were collected for valve sizing and vascular access assessment, respectively. Valve sizing was assessed by an expert interventional cardiologist using initial manual multiplanar reformatting to identify the annular plane, followed by standard measurements, including annular area, perimeter, sinus height, eccentricity, and minimum and maximum diameters [16-18]. The study was approved by the institutional review committee, and informed consent was obtained before patient enrollment. Baseline clinical characteristics of the patient population included age, sex, body surface area (BSA), body mass index (BMI), Society of Thoracic Surgeons Predicted Risk of Mortality (STS-PROM), diabetes, New York Heart Association (NYHA) functional class, and bicuspid aortic stenosis, as summarized in Table 1.

**Table 1**
Baseline characteristics of the patient population (n = 109).

| | |
|---|---|
| Age, years | 80 ± 5.5 |
| Female sex | 50.1 % |
| BSA, $m^2$ | 1.80 ± 0.39 |
| BMI, $kg/m^2$ | 28.1 ± 6.0 |
| STS PROM, % | 4.6 ± 4.2 |
| Diabetes | 33.3 % |
| NYHA class 3 or 4 | 70.5 % |
| Bicuspid aortic stenosis | 5.3 % |

### 2.2 AI Workflow

#### 2.2.1 Automated Annotations

TAVI-TEC is an artificial intelligence-driven software that implements a fully automated pipeline for analysis of pre-procedural CT images. Figure 1 shows the overall pipeline for automatic pre-TAVI assessment, beginning with DICOM image conversion and proceeding through anatomical segmentation, landmark detection, centerline extraction, annotations, and prosthesis size prediction using machine learning. CT images were first converted from DICOM to Neuroimaging Informatics Technology Initiative (NIfTI) format to allow standardized handling of three-dimensional imaging volumes. The converted images were then processed using dedicated deep learning-based segmentation modules designed to identify the main cardiac and vascular structures (ie, the left ventricle and aorta) [19]. After segmentation, the output masks were binarized using a threshold value of 0, whereby all voxels greater than 0 were retained as part of the structures. Coronary arteries were segmented separately, and their masks were binarized and refined by retaining the two largest connected components, corresponding to the left coronary artery (LCA) and right coronary artery (RCA). Segmentation masks were then used to automatically extract the centerline by skeletonization with 26-neighborhood connectivity. Annulus location was determined along the extracted centerline as the transition point at which the trajectory exited the left ventricular mask. This approach enabled multiplanar reformatting and identification of the annular plane using the centerline tangent at the annular point. The annular plane was further refined by translating the plane along the local centerline direction and applying small angular rotations around two orthogonal axes. The final annular plane was identified as the valid plane with the minimum cross-sectional area. The sinuses and sinotubular junction (STJ) were subsequently identified from the cross-sectional radius profile computed along the centerline starting from the annulus. Sinus diameters were measured on a cross-sectional plane located at half sinus height and oriented orthogonally to the local centerline tangent. Aortic calcifications were segmented within a dilated aorta-left ventricle region of interest (ROI) using an intensity-based threshold. The threshold was defined as the 99.5th percentile of Hounsfield unit (HU) values within the ROI, with voxels above this threshold classified as calcifications.

For annotation, the annular contour of the segmented aortic root was defined using the convex hull of the projected annular point, from which area, perimeter, major and minor axes, and eccentricity were computed. Landmark heights were calculated as centerline distances from the annulus to the sinus mid-height, STJ, and coronary ostia projections, whereas calcium burden was quantified as the total volume of segmented calcification voxels. Curved planar reformation (CPR) was generated by resampling the CT volume along the smoothed centerline to obtain a longitudinal reformatted view of the CTA scan.

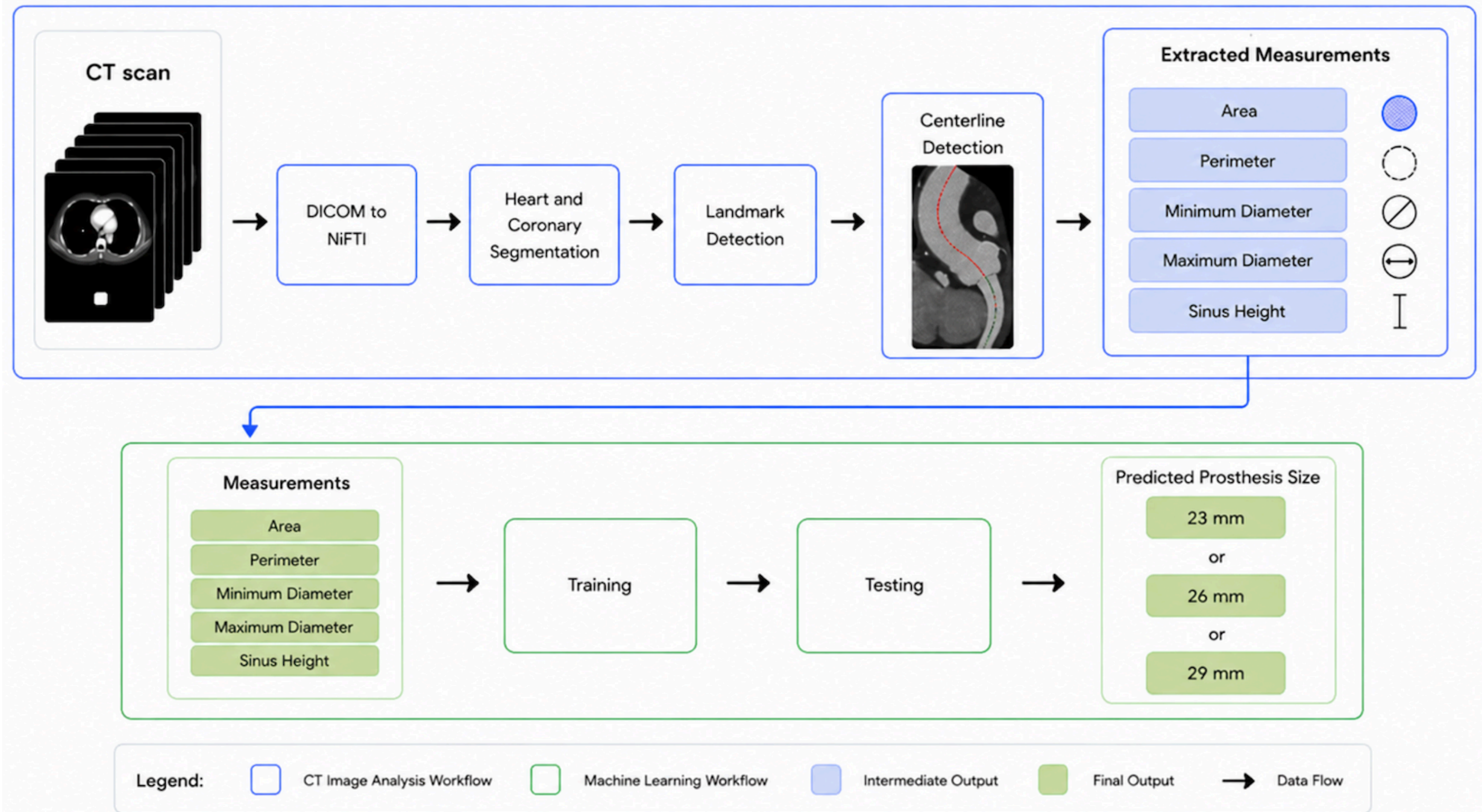

**Figure 1.**
Pipeline of the computational framework for automated annotations and valve-size prediction.

### 2.2.2 Vascular Access Assessment

Full-body CTA scans were used to segment the aorta, iliac arteries, and associated calcified plaques. The analysis followed an approach similar to that used for the aortic root, as this was based on automatic vessel segmentation and centerline extraction. Qualitative vascular access assessment was provided through three-dimensional visualization of vessel diameter maps, calcification distribution, and lumen narrowing caused by calcified plaques. In addition, TAVI-TEC generated CPR views of the iliac and femoral arteries to automatically visualize calcifications along the vessel centerline and evaluate the residual lumen area available for delivery of the crimped device.

### 2.2.3 Machine Learning

Valve-size prediction was performed using a supervised multiclass multilayer perceptron classifier. Five input features, corresponding to annular measurements, were extracted for each patient and normalized to the range [-1, 1]. Target classes corresponded to the implanted S3 valve sizes: 23, 26, and 29 mm. The network consisted of one hidden layer with 10 sigmoid neurons and a three-class output layer. Weights were initialized using the Nguyen-Widrow method, and training was performed with a limited-memory Broyden-Fletcher-Goldfarb-Shanno (LBFGS) optimizer using cross-entropy loss with L2 regularization. Model performance was evaluated using stratified two-fold cross-validation while preserving class distribution across folds. The best-performing fold was selected for final evaluation. Classification performance was assessed using receiver operating characteristic (ROC) analysis with area under the curve (AUC) calculation. Confusion matrices were used to visualize class predictions, and accuracy, precision, recall, and F1-score were calculated for each prosthesis-size class. Misclassifications were further analyzed as false-positive and false-negative predictions for each prosthesis-size class. Agreement between predicted and implanted prosthesis size was categorized as smaller-size prediction, exact agreement, or larger-size prediction.

**2.3 Statistical Analysis**

Statistical analysis was performed using STATA statistical software (StataCorp, USA). Data are presented as mean +/- standard deviation (SD). Agreement between TAVI-TEC measurements and clinician-derived measurements was evaluated using the concordance correlation coefficient (CCC), coefficient of variation (CV), intraclass correlation coefficient (ICC), and Bland-Altman analysis. Linear regression analysis was performed to assess the relationship between automated and manual measurements, with the identity line used as a reference for consistency. Bland-Altman analysis was used to identify systematic bias and determine the limits of agreement between measurement methods.

**2.4 Software integration with DICOM Vision**

The automated pipeline was integrated into DICOM Vision as a web-based workflow module. CTA studies can be loaded into the viewer, processed by a server-side AI pipeline, and returned to the user interface as structured measurements, segmentation overlays, CPR views, and three-dimensional visualizations. The integration allows users to inspect the generated outputs within the same environment used for image visualization, reducing the need to move data across separate software tools. This architecture is intended to support future integration with standard imaging workflows, including PACS/DICOM-based environments and structured reporting.

## 3. RESULTS

Figure 2 shows screenshots of the TAVI-TEC application implemented in the web-based DICOM_Vision medical imaging software developed by D/Vision Lab. Figure 3 illustrates the output of the TAVI-TEC application, including three-dimensional visualization of the aortic root and corresponding pre-procedural TAVI measurements. These measurements include annular perimeter and area, highlighted in red, with green and blue lines representing maximum (Dmax) and minimum (Dmin) annular diameters, respectively. Markers indicate sinus and STJ heights, as well as projections of the coronary artery ostia onto the vessel centerline. Sagittal and axial CPR views along the vessel centerline enable assessment of calcification patterns and visualization of the annular cross-section at the measurement plane. Table 2 summarizes sinus diameter, LCA height, RCA height, and STJ height for the study cohort, as obtained using TAVI-TEC. Computational time ranged from 2.2 to 6.4 min using a Tesla T4 graphics processing unit (GPU).

**Table 2. Average aortic root height measurements (mm) computed by TAVI-TEC.**

| Sinus height | LCA height | RCA height | STJ height |
|---|---|---|---|
| 22.3 ± 4.2 | 17.5 ± 4.2 | 22.0 ± 4.2 | 26.0 ± 4.2 |

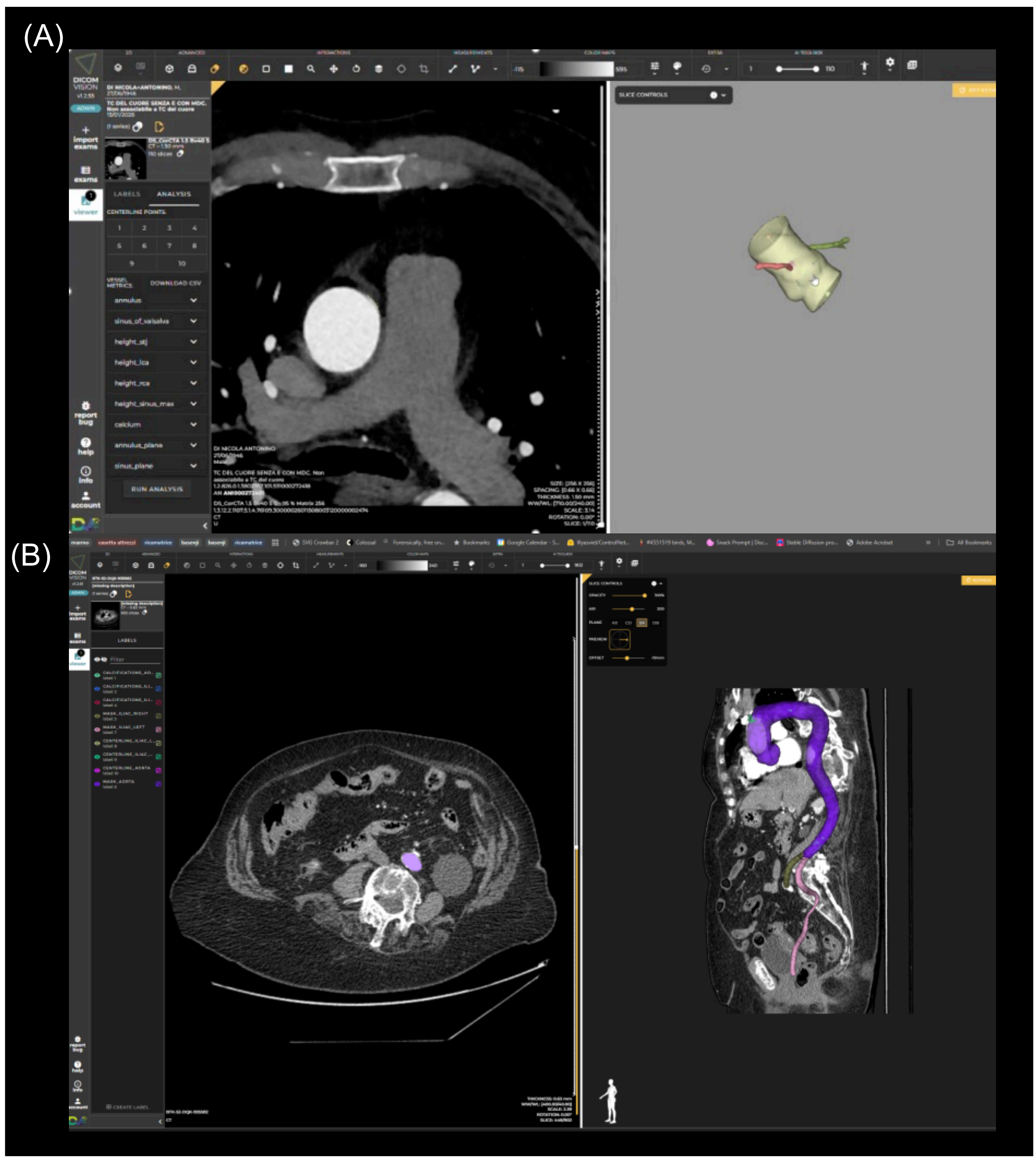

**Figure 2.** Screenshots of the TAVI-TEC framework integrated into the DICOM_Vision web-based medical imaging platform developed by D/Vision Lab. (A) Axial view of the aorta and three-dimensional view of the segmented aortic root with annotations in the right panel of the DICOM Vision software. (B) Full-body scan with three-dimensional view of the segmented aorta for vascular access assessment.

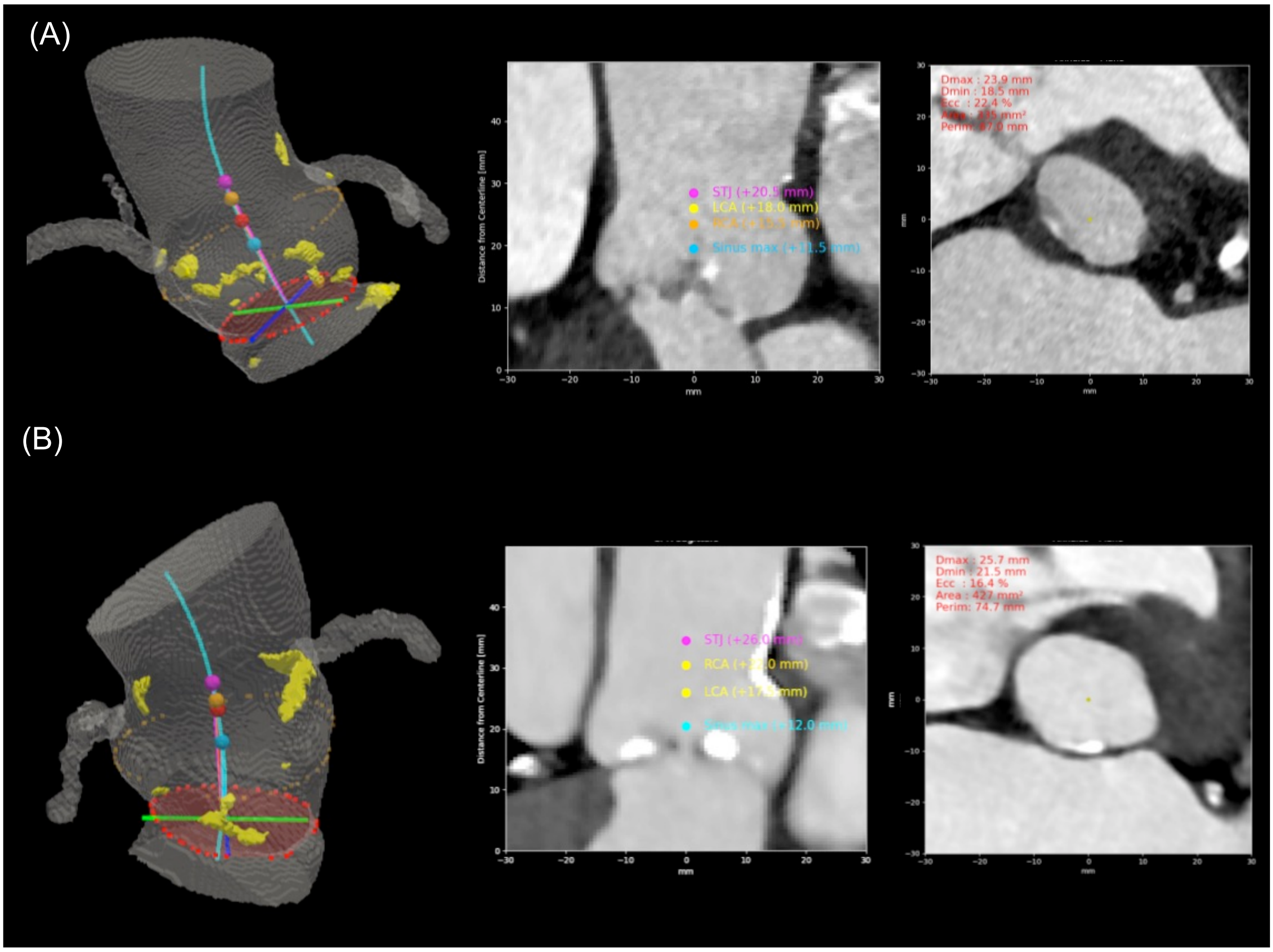

**Figure 3.** Software output showing the three-dimensional model of the aortic root (gray) and calcifications (yellow), with labels for two representative TAVI patients (A and B), together with CPR-related sagittal and axial views. The red plane indicates the annulus, the yellow plane indicates the sinus of Valsalva, the blue dot indicates sinus height, the red dot indicates the right coronary artery, the yellow dot indicates the left coronary artery, and the pink dot indicates STJ height.

Assessment of vascular access for transfemoral TAVI procedures was facilitated by visualization of calcification patterns along the aorta, as illustrated in Figure 4A. In addition, CPR views of the left and right iliac arteries enabled detailed evaluation of vessel narrowing. The TAVI-TEC application automatically detected calcifications and provided cross-sectional views of the iliac arteries, with the number and location of views adapted to the automatically extracted numbers of calcific plaque burden in the investigated patient. Vascular access analysis was further supported by color-coded maps representing lumen reduction induced by thick calcific plaques (Figure 4B) and vessel diameter distribution along the centerline (Figure 4C).

Table 3 summarizes the agreement between clinician-derived measurements and TAVI-TEC outputs. Overall, a high level of agreement was observed for annular area and perimeter, with strong correlations (R2 = 0.881 and 0.854, respectively) and regression coefficients close to unity (beta = 0.972 and 0.968). These parameters also demonstrate excellent concordance, as indicated by high CCC values (0.934 and 0.909) and ICC values (ICC individual = 0.935 and 0.909; ICC one-way = 0.966 and 0.952). Similarly, maximum (Dmax) and minimum (Dmin) annular diameters showed good agreement, with R2 values of 0.649 and 0.752 and regression coefficients of 0.853 and 0.901, respectively. Concordance and reliability remained acceptable for the annulus diametes, with CCC values of 0.794 for Dmax and 0.823 for Dmin and ICC individual values of 0.794 and 0.820, respectively.

**Table 3. Agreement between TAVI-TEC and clinician-derived measurements.**

| Parameter | R2 | Beta | Beta SE | CCC | CV | ICC individual | ICC one-way |
|---|---|---|---|---|---|---|---|
| Area | 0.881 | 0.972 | 0.038 | 0.934 | 6.132 | 0.935 | 0.966 |
| Perimeter | 0.854 | 0.968 | 0.043 | 0.909 | 3.419 | 0.909 | 0.952 |
| Dmax | 0.649 | 0.853 | 0.067 | 0.794 | 6.288 | 0.794 | 0.885 |
| Dmin | 0.752 | 0.901 | 0.055 | 0.823 | 4.648 | 0.820 | 0.901 |

**Note**: R2 = coefficient of determination; beta = regression coefficient; beta SE = standard error of the regression coefficient; CCC = concordance correlation coefficient; CV = coefficient of variation; ICC = intraclass correlation coefficient; ICC individual = intraclass correlation coefficient for individual measurements; ICC one-way = intraclass correlation coefficient for one-way random-effects models.

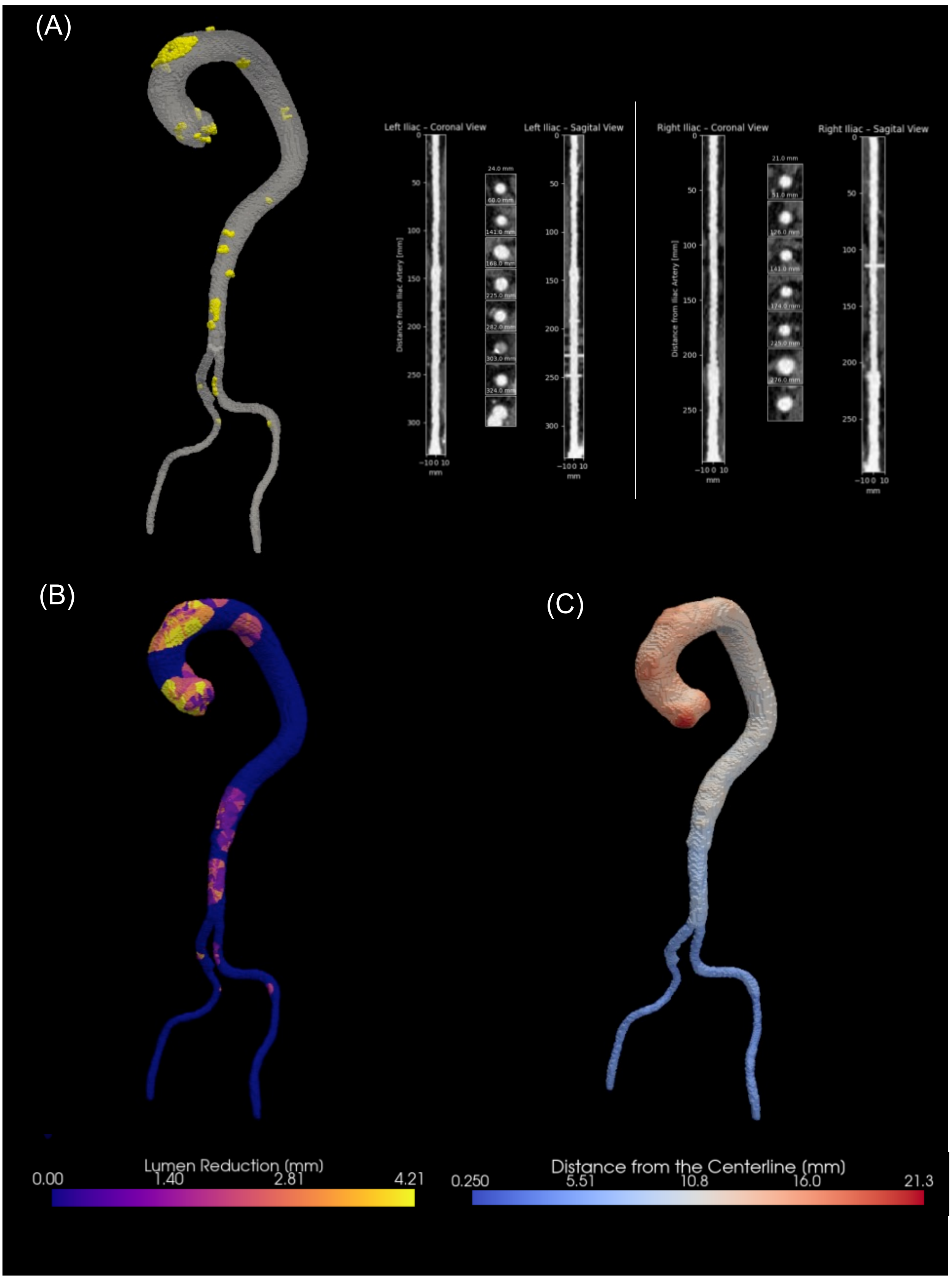

**Figure 4.** (A) Three-dimensional view of the aorta (gray) with calcifications (yellow) for vascular access evaluation, together with CPR views of the right and left iliac arteries and cross-sectional views of plaques. (B) Distribution of lumen reduction due to calcification thickness. (C) Distribution of distance from the centerline.

For all pre-procedural parameters (Figures 5 and 6), agreement between methods was further evaluated using Bland-Altman analysis to quantify systematic bias and limits of agreement, together with linear regression to assess the relationship between measurements. For annular area and perimeter, Bland-Altman analysis indicated a small negative bias (-6.7 for area and -1.16 for perimeter), with most differences distributed around the mean and within the limits of agreement. No evident proportional bias was observed because differences were relatively homogeneously distributed across the measurement range. Linear regression demonstrated a strong linear relationship between the two methods, with data points closely aligned along the identity line, indicating high consistency and positive correlation. Comparable findings were observed for diameter measurements; although, the differences between manual and automatic measurements were slightly higher (0.34 for the major diameter and -1.6 for the minor diameter) than annulus area and perimeter.

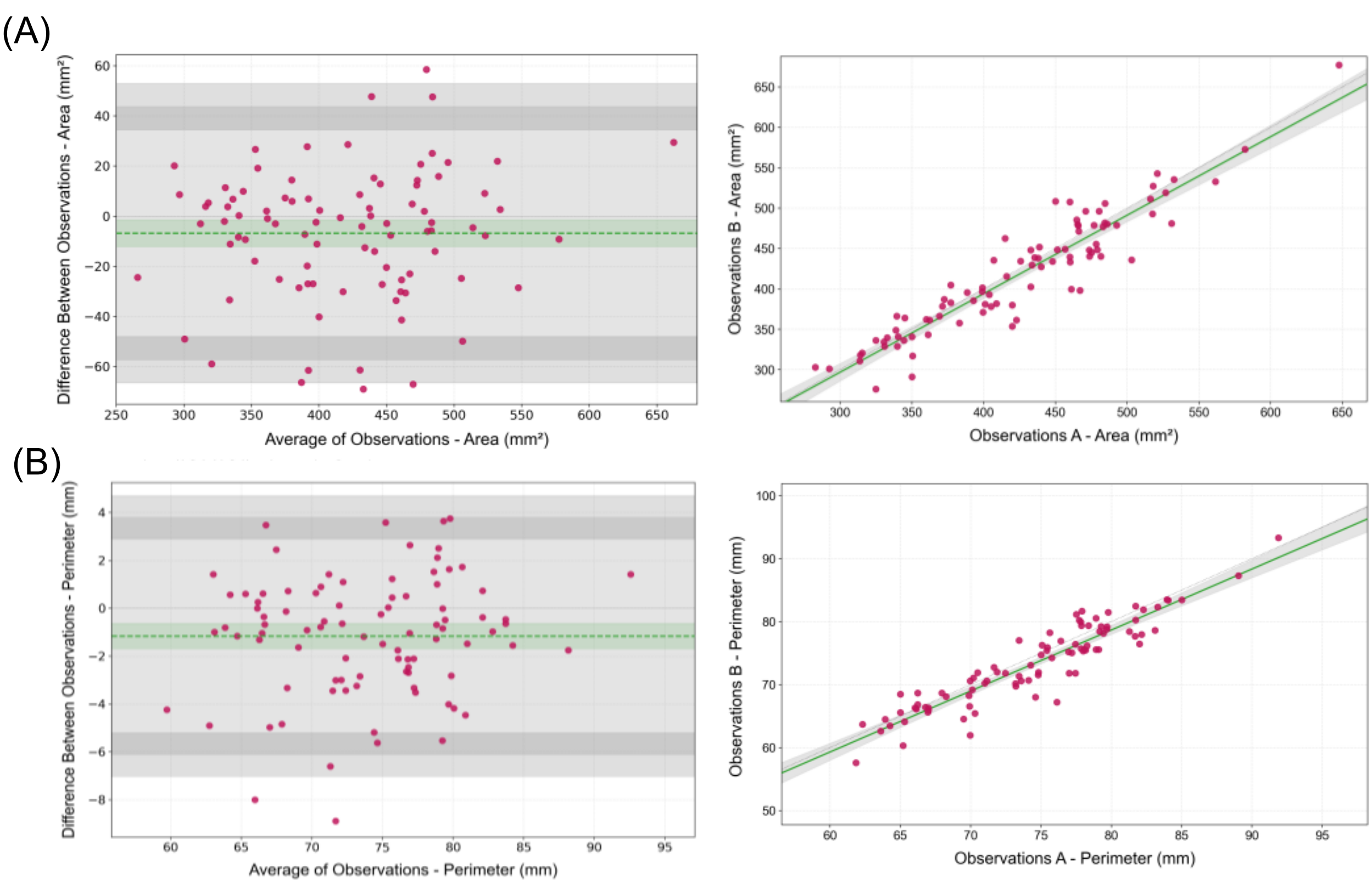

**Figure 5.** Bland-Altman plots and regression curves for (A) annular area and (B) perimeter measured by clinicians and the TAVI-TEC application.

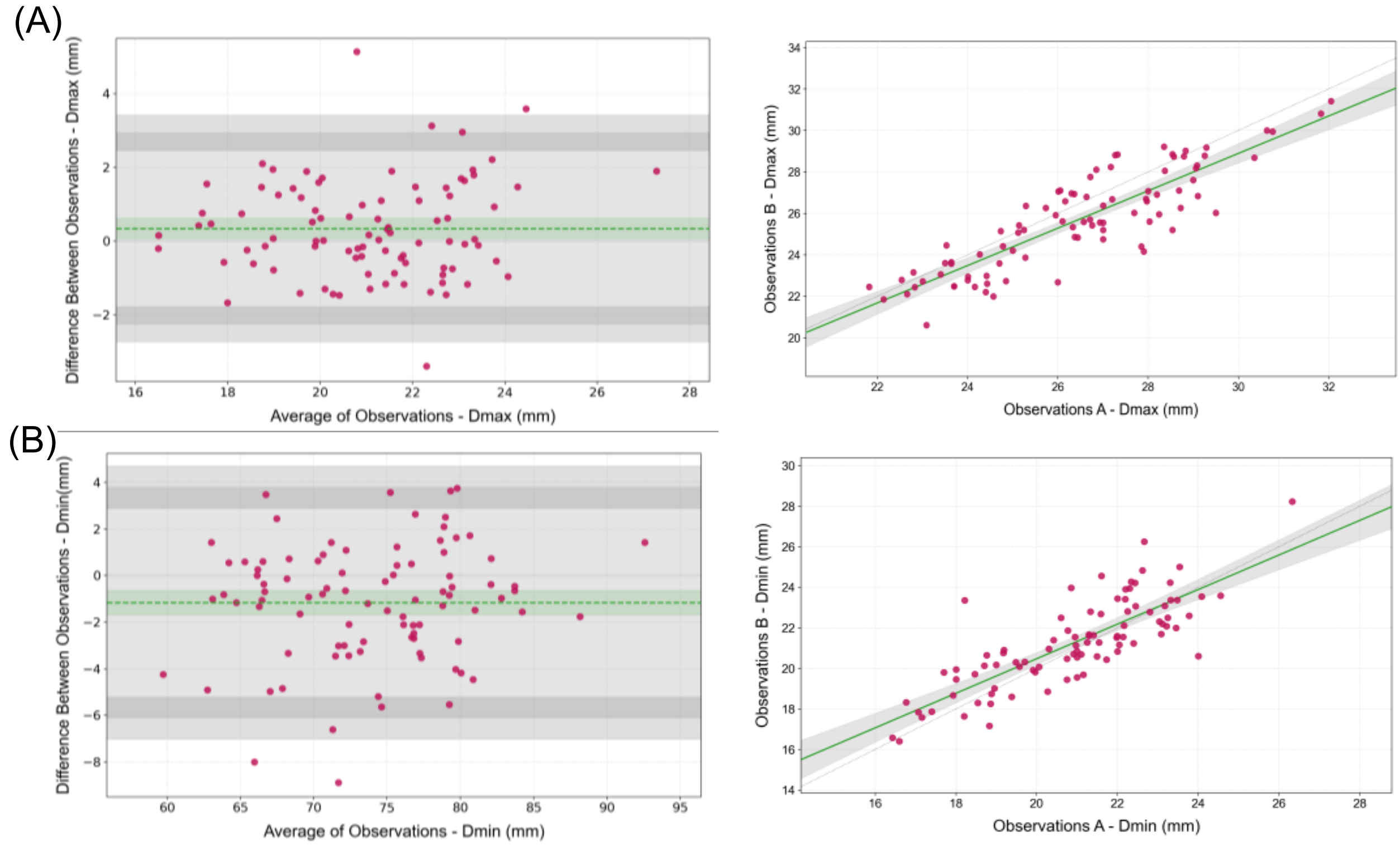

**Figure 6.** Bland-Altman plots and regression curves for (A) Dmax and (B) Dmin measured by clinicians and the TAVI-TEC application.

The multilayer perceptron achieved an overall accuracy of 82% in the best-performing cross-validation fold. For the 23-mm SAPIEN 3 Ultra valve, the model demonstrated strong performance, with precision of 0.83, recall of 0.86, and F1-score of 0.84. Performance for the 26-mm valve was moderate, with precision of 0.75, recall of 0.71, and F1-score of 0.73. For the 29-mm valve, precision, recall and F1-score were all 0.91 indicating balanced and robust classification performance for this class. Confusion matrix analysis indicated that most misclassifications occurred between adjacent prosthesis sizes (Figure 7A). ROC analysis demonstrated high discriminative ability across all classes, with AUC values of 0.943 for the 23-mm valve, 0.865 for the 26-mm valve, and 0.993 for the 29-mm valve (Figure 7B). Undersizing, defined as prediction of a smaller valve, occurred in 10% of cases, whereas oversizing, defined as prediction of a larger valve, was observed in 8% of cases (Figure 7C).

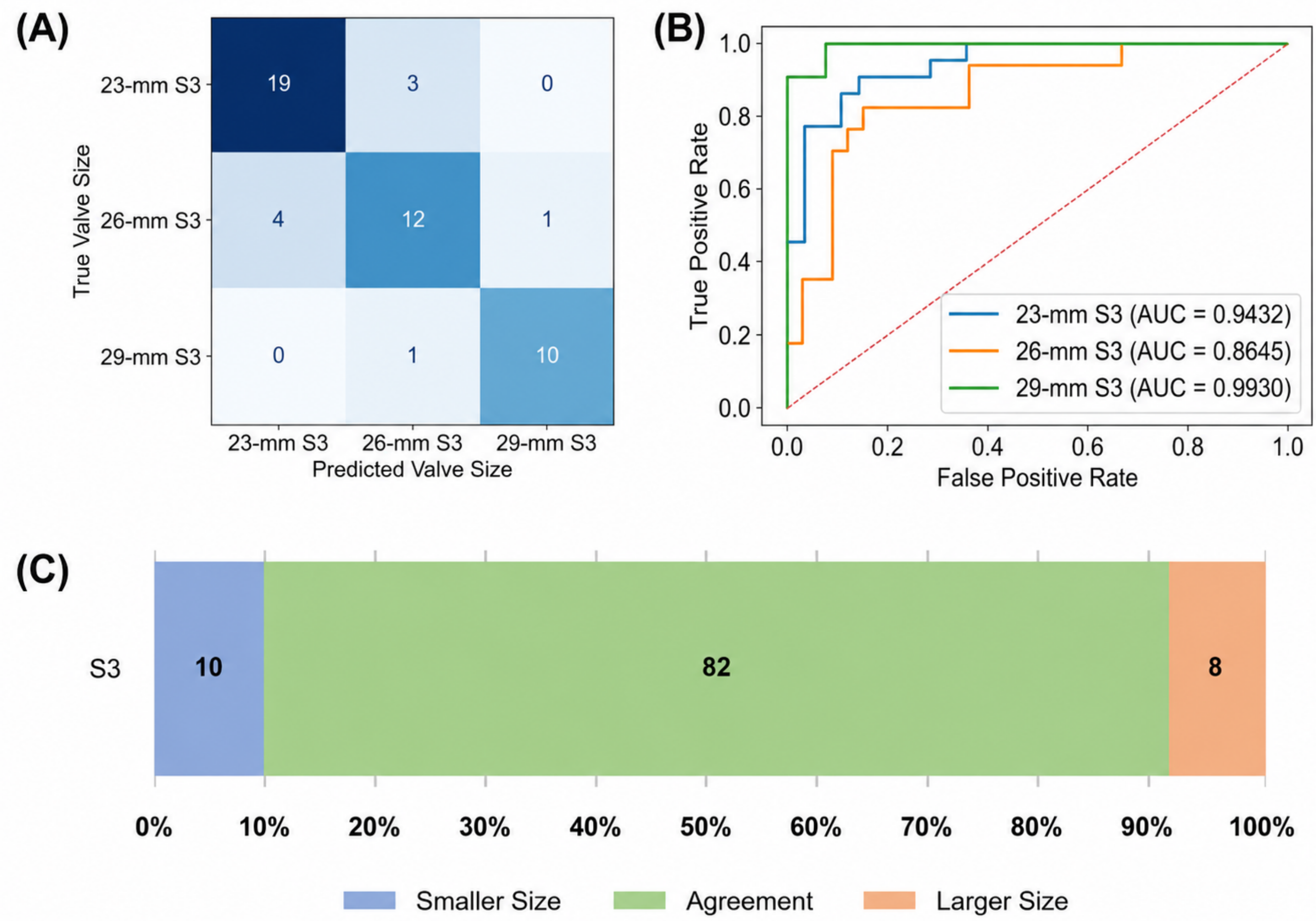

**Figure 7.** (A) Confusion matrix for the multiclass multilayer perceptron neural network across SAPIEN 3 Ultra device sizes. (B) ROC curve for each device size. (C) Device-size agreement, defined as concordance between predicted and clinically selected sizes, with mismatches categorized as undersizing or oversizing**.**

## 4. DISCUSSION

The present study introduces a fully automated AI-based tool for extracting fundamental pre-TAVI measurements from routine CT imaging and supporting device-sizing evaluation based on predicted annular dimensions. The automated nature of the software offers important advantages in routine clinical practice, as it enables a reduction in planning time, optimization of hospital resources, and streamlining of the pre-procedural work-up of patients. These aspects may become increasingly important as the number of TAVI procedures continues to rise, driven by the expansion of indications toward lower risk patients. TAVI-TEC performs segmentation of the aortic root and calcifications using deep learning algorithms, followed by detection of anatomical landmarks from which measurements are derived. The tool is accessible through the web-based DICOM_Vision medical imaging platform developed by D/Vision Lab, facilitating integration into the clinical workflow. Computation required few minutes (≤ 7 min in all cases) on a GPU architecture to extract anatomical structures and associated measurements. Systematic comparison with manual measurements obtained by an expert clinician in a cohort of 109 patients treated with the S3U device demonstrated strong agreement, with intraclass correlation coefficients ranging from 0.82 to 0.93. Vascular access assessment was achieved through color-coded maps of lumen reduction and aortic diameter, complemented by automatically generated CPR views in regions exhibiting iliac artery narrowing. In addition, AI-based device-size prediction achieved 82% agreement with implanted device sizes. Overall, TAVI-TEC represents a robust solution for pre-procedural imaging analysis in TAVI, enabling fully automated and standardized quantification of anatomically relevant features in aortic stenosis.

Strong agreement between manual and automated measurements at the aortic annulus was observed. The primary source of discrepancy was likely related to differences in annular-plane identification, which may vary among observers. Although Toggweiler et al. [14] reported correlation coefficients exceeding 0.95, measurements obtained with 3mensio relied on a semi-automated approach in which nadir positions were suggested to the user before parameter evaluation. Consequently, measurements were performed on the same annular plane by both the user and the software, resulting in negligible discrepancies. Saitta et al. [12] also reported good agreement between manual and automated measurements using a three-dimensional U-Net framework for segmentation combined with differential geometry for landmark annotation, with a computational time of less than 45 s using a high-end GPU system (Nvidia A100). In a large cohort of 200 TAVI patients, the TAVI-PREP [15] method demonstrated superior performance in terms of correlation coefficients for annular measurements, with the exception of coronary height when compared with the approach reported by Astudillo et al [20]. TAVI-PREP defines the annular plane based on the three nadir points. In contrast, TAVI-TEC initially defines the annular plane using the three nadirs, followed by systematic proximal and distal adjustments of the annulus planes thanks to rotational perturbations of up to 5 degrees relative to the vessel centerline. The plane that minimizes cross-sectional vessel area is selected for annotation. This strategy ensures that measurements are obtained under a conservative, worst-case configuration while maintaining a standardized methodology and reducing operator-dependent variability.

Compared with other AI-based pre-TAVI tools [14, 15, 20], TAVI-TEC provides clinical-grade visualization of calcifications to support vascular access assessment. Minimization of access-related complications is as critical as accurate valve sizing. This goal is addressed through automatically generated CPR views of narrowed iliac arteries, complemented by cross-sectional axial views of plaques distal to the aortic bifurcation, tailored to the extent of calcific burden in each patient. Evaluation of transfemoral access is further supported by three-dimensional visualization of the entire aorta, with interactive manipulation and color-coded maps representing lumen reduction due to calcifications and aortic diameter distribution. In addition, TAVI-TEC projects coronary ostia onto the vessel centerline, facilitating an alternative spatial interpretation of coronary height and associated risk compared with conventional Euclidean distance-based methods, such as that of TAVI-PREP. Unlike several research-oriented prototypes, TAVI-TEC was implemented within the DICOM Vision web-based medical imaging platform, allowing AI-generated measurements, segmentations, CPR views, and 3D visualizations to be reviewed in a single imaging environment. This integration represents a relevant step toward translation of automated pre-TAVI planning from algorithmic development to workflow-oriented software deployment.

The multilayer perceptron demonstrated robustness for multiclass classification of S3U sizing. In a previous study from our group [21], this model exhibited superior predictive performance compared with support vector machines and k-nearest-neighbor classifiers. However, that model was based on geometric features derived from statistical shape modeling and did not support prediction of the 29-mm SAPIEN 3 Ultra valve. In contrast, the TAVI-TEC neural architecture was refined to enable predictions based on a set of clinically relevant aortic root measurements. The model achieved AUC values of 0.943 for the 23-mm device, 0.865 for the 26-mm device, and 0.993 for the 29-mm device. The highest misclassification rate was observed for the 29-mm valve, likely because this device size was underrepresented in the study population as large S3U valves are implanted less frequently. For clinical validation of sizing classification, the tool correctly matched the implanted valve size in 82% of cases. This agreement rate was slightly lower than that reported for 3mensio (87% for Accurate and 88% for Evolut and SAPIEN 3) and by Astudillo et al. [20] (86% for SAPIEN 3 and 89% for Evolut), but substantially higher than values reported by Kruger et al. [13] (61%-64%; device not specified). In cases of disagreement with clinical selection, TAVI-TEC tended to recommend a smaller valve size. Error analysis in collaboration with clinicians indicated that discrepant cases were predominantly borderline between the 23-mm and 26-mm S3U sizes.

The primary limitation of this study is the single-center design and evaluation of a single device platform. In addition, several measurements relevant to comprehensive TAVI planning were not included, such as sinus diameter. The accuracy of the deep learning-based segmentation was not formally evaluated and may contribute to variability in the derived annotations. The proposed AI-based approach should therefore be considered as a support tool for clinical decisions since clinical judgment by experienced physician teams remains essential.

## 5. CONCLUSION

The TAVI-TEC methodology provides a fully automated, AI-based framework for pre-procedural TAVI planning using routine CTA imaging. The proposed pipeline permitted a rapid segmentation of relevant cardiovascular structures, extraction of key annular and aortic root measurements, visualization of calcifications and vascular access anatomy as well as prediction of S3U sizing. Compared with clinician-derived measurements, TAVI-TEC demonstrated strong agreement for annular area and perimeter and acceptable agreement for annular diameters with differences likely caused by the annular plane identification. Integration into a web-based DICOM viewer further supports the potential implementation in routine clinical workflows. Though the application may be extended to include additional measurements validated in large cohort, TAVI-TEC has the potential to reduce operator-dependent variability, streamline preoperative planning, and support Heart Team decision-making during preoperative TAVI assessment.

## Declarations

### Ethics Approval and Consent to Participate

This study was approved by the IRCCS ISMETT Ethics Committee (approval no. IRRB04/04). All participants provided written informed consent before enrollment.

### Consent for Publication

All authors were involved in the study and manuscript preparation, and the originality of their contributions can be confirmed by members of ISMETT and the University of Palermo. All authors approved submission of the manuscript.

### Availability of Data and Materials

The datasets generated and/or analyzed during the current study are not publicly available because of ethical restrictions but are available from the corresponding author upon reasonable request.

### Competing Interests

The authors declare that they have no known competing financial interests or personal relationships that could have appeared to influence the work reported in this paper.

### Funding

This project received funding from the European Commission - NextGenerationEU - PNRR M6 - C2 - Investimento 1.3: Sviluppo di tecnologie e percorsi innovativi per la salute - Iniziativa PNC0000003, titled "ANTHEM: AdvaNced Technologies for Human-centrEd Medicine" - CUP B53C22006700001.